\documentclass{article}

\usepackage[preprint]{nips_2017}

\usepackage[utf8]{inputenc} 
\usepackage[T1]{fontenc}    
\usepackage{hyperref}       
\usepackage{url}            
\usepackage{booktabs}       
\usepackage{amsfonts}       
\usepackage{nicefrac}       
\usepackage{microtype}      
\usepackage{microtype}
\usepackage{subfigure}
\usepackage{amssymb}
\usepackage{amsmath}
\usepackage{multirow}
\usepackage{graphicx}

\DeclareMathOperator{\disc}{disc}

\title{Theoretical Guarantees of Transfer Learning}

\author{
  Zirui Wang \\
  Language Technologies Institute\\
  Carnegie Mellon University\\
  Pittsburgh, PA 15213 \\
  \texttt{ziruiw@cs.cmu.edu} \\
}

\begin{document}

\maketitle

\begin{abstract}
  Transfer learning has been proven effective when within-target labeled data is scarce. A lot of works have developed successful algorithms and empirically observed positive transfer effect that improves target generalization error using source knowledge. However, theoretical analysis of transfer learning is more challenging due to the nature of the problem and thus is less studied. In this report, we do a survey of theoretical works in transfer learning and summarize key theoretical guarantees that prove the effectiveness of transfer learning. The theoretical bounds are derived using model complexity and learning algorithm stability. As we should see, these works exhibit a trade-off between tight bounds and restrictive assumptions. Moreover, we also prove a new generalization bound for the multi-source transfer learning problem using the VC-theory, which is more informative than the one proved in previous work.
\end{abstract}

\section{Introduction}
Traditional supervised machine learning methods share the
common assumption that training data and test data are
drawn from the same underlying distribution. Typically, they
also require sufficient labeled training instances to construct
accurate models. In practice, however, one often can only
obtain limited labeled training instances. Inspired by human
beings’ ability to transfer previously learned knowledge
to a related task, transfer learning [1] addresses the challenge of data scarcity in the target domain by utilizing labeled data from other related source domain(s).

Plenty of research has been done on the general transfer learning problem [2, 3, 4]. These methods typically involve selecting a subset of source samples and/or features to transfer knowledge from the source to the target. While most methods are developed on top of heuristic, good empirical results have been achieved on a wide range of applications in areas such as sentiment analysis [5], computer vision [6], cross-lingual natural language processing [7], and urban computing [8].

A goal in machine learning is typically to minimize the expected risk. Given that we usually cannot directly measure the true risk, the alternative goal is to minimize the empirical risk. A type of learning theory known as the generalization theory aims to explain why minimizing this empirical risk is a sensible approach to minimize the true risk by analyzing generalization bounds. While many efforts have been dedicated in such theory for analyzing generalization gap in machine learning models [9], such works are usually not directly applicable to the transfer learning settings. This is due to the fact that traditional learning theory such as PAC framework is built on the assumption that functions are estimated using random samples (usually iid) drawn from the same target distribution. However, transfer learning models are trained using labeled samples drawn from the source distribution, which is different from the target distribution that the true risk measured over. Therefore, analyzing the generalization bound for transfer learning is a challenging problem.

There are currently several different approaches of analyzing the generalization gap in the machine learning community. Among which the most famous one is the \textit{model complexity} approach that measures the generalization bound by exploiting the complexity of hypotheses set such as Vapnik–Chervonenkis (VC) dimension [10] and Rademacher complexity [9]. Another approach [11] tries to utilize the \textit{stability} of the supervised learning algorithm with respect to datasets. The stability is a measure of how much changing a data point in the training samples can change the algorithm output. Both of these two approaches have been applied to the analysis of generalization bounds of transfer learning algorithms.

The model complexity approach is applied in a more specific setting of transfer learning in which the marginal distributions over the source and the target domains are different while the conditional distributions are assumed to be the same (or at least similar). This is known as the problem of Domain Adaptation (DA) [12], where a successful scheme typically utilizes large unlabeled samples from both domains to adapt a source hypothesis to the target domain. The earliest theoretical work of transfer learning (domain adaptation) is [13]. The work proves a VC-bound on the expected risk of a target hypothesis, when the target hypothesis minimizes a convex combination of the empirical source and target risks. The work is further extended in [14] and [15], with applications to domain adaptation using multiple sources. A similar setting was also examined by Mansour et al. [16], whom instead derived the generalization bound using Rademacher complexity but got similar results with more analysis under various scenarios.

At the other end of the spectrum, another approach using algorithm stability is examined under a different transfer learning setting known as the Hypothesis Transfer Learning (HTL) [17]. Notice that in other settings such domain adaptation, large amount of unlabeled data are usually required for both domains to estimate domain divergence and the importance weight for the source domain (as in the case of convex combination in [13]). Hypothesis transfer learning relaxed this assumption by only assuming that source hypotheses trained on a source domain are available. Therefore, it does not require direct access to the source training samples, nor any knowledge about the relatedness of the source and target distributions. As a result, the theoretical analysis of hypothesis transfer learning surpasses the distribution gap in common transfer learning problems. The first theoretical work was presented in [18], in which a generalization bound using hypothesis stability defined in [11] was derived for hypothesis transfer learning. A side effect is that they showed that hypothesis transfer learning is resistant to negative transfer [19], a phenomena when transferring from the source domain actually hurts the performance in the target domain. A study of theoretical bounds of hypothesis transfer learning in multi-source settings is presented in [20]. And a more recent work [21] using transformation functions has shown that hypothesis transfer learning enjoys faster convergence rates of excess risks for Kernel methods.

\section{Notations \& Definitions}

Here, we first introduce a few notations and definitions. First we define the generalization theory in the general machine learning problem. Let $R[f_{A(S)}]$ denotes the expected risk of the output function $f_{A(S)}$ of a supervised learning algorithm $A$ applied to training sample $S$, and let $\hat{R}[f_{A(S)}]$ be the empirical training risk. Since learning algorithms output optimal functions by minimizing this empirical training risk, $f_{A(S)}$ has a dependency on the data used to estimate the risk and thus $\hat{R}[f_{A(S)}]$ is biased. To analyze the generalization gap, defined as:
$$ R[f_{A(S)}] - \hat{R}[f_{A(S)}] $$
several approaches have been proposed, such as model complexity, stability, robustness and flat minima. In this paper, we show the theory of transfer learning using model complexity and stability.

Model complexity approach overcomes this dependence using the following trick:
$$ R[f_{A(S)}] - \hat{R}[f_{A(S)}] \leq \sup_{f \in F}  R[f] - \hat{R}[f]$$
And by introducing a quantity such as VC dimension or Rademacher complexity to characterize $F$, we can derive a union bound for such approach.

As we should see, the complexity bound for transfer learning (domain adaptation) is a direct extension of such strategy by introducing additional distribution divergence measures.

We first formalize domain adaptation for binary classification similar to [15]. Let $\mathcal{X}$ be the input space, $\mathcal{D}$ denotes a distribution on $\mathcal{X}$ and $f: \mathcal{X} \rightarrow [0, 1]$ denotes the labelling function. For the basic case of a single source, the source domain is denoted as $<\mathcal{D}_S, f_S>$ and the target domain is denoted as $<\mathcal{D}_T, f_T>$.

For any hypothesis $h$, we use a 0-1 loss function and denote the expected risk according to distribution $\mathcal{D}$ and labelling function $f$ as:
$$ \epsilon_{\mathcal{D}}(h, f) = \mathbb{E}_{x \sim \mathcal{D}}[|h(x)-f(x)|]$$
For simplicity, we use the shorthand $\epsilon_S(h)=\epsilon_{\mathcal{D}_S}(h, f_S)$ and write the empirical risk as $\hat{\epsilon}_S(h)$. Notations for the target domain are parallel. Notice that these notations are equivalent to the risk notations given above.

To measure the distance between two distributions, we use the following definition of distribution divergence:

\textbf{Definition 1} {\itshape For a hypothesis space $\mathcal{H}$ for instance space $\mathcal{X}$, the symmetric difference hypothesis space $\mathcal{H}\Delta\mathcal{H}$ is the set of hypotheses defined as}
\begin{equation}\label{eq:1}
\mathcal{H}\Delta\mathcal{H} = \{ h(x)\oplus h'(x): h, h' \in \mathcal{H} \},
\end{equation}
{\itshape where $\oplus$ is the XOR function. Then the $\mathcal{H}\Delta\mathcal{H}$-divergence between any two distributions $D$ and $D'$ is defined as}
\begin{equation}\label{eq:2}
d_{\mathcal{H}\Delta\mathcal{H}}(\mathcal{D}, \mathcal{D}') \triangleq 2 \mathop{\sup}\limits_{A\in\mathcal{A}_{\mathcal{H}\Delta\mathcal{H}}}\Big|Pr_{\mathcal{D}} [A]-Pr_{\mathcal{D}'} [A]\Big|,
\end{equation}
{\itshape where $\mathcal{A}_{\mathcal{H}\Delta\mathcal{H}}$ is the measurable set of subsets of $\mathcal{X}$ that are the support of some hypothesis in $\mathcal{H}\Delta\mathcal{H}$.}

The above definition of divergence measures the maximal discrepancy between two distributions given a fixed hypothesis class. It is also known as the $\mathcal{A}$-distance in [13] and it is also equivalent to the definition of total variational divergence in the binary case. For our purposes, this distance measure has an important advantage over other means of comparing distributions such as $L_1$ distance, KL divergence or $\chi^2$ divergence. It's that we can compute the $\mathcal{H}\Delta\mathcal{H}$-divergence from finite unlabeled samples of the distributions $\mathcal{D}$ and $\mathcal{D}'$ when $\mathcal{H}$ has finite VC dimension. Furthermore, we can compute a finite-sample approximation by finding a classifier $h\in\mathcal{H}$ that maximally discriminates between (unlabeled) instances from $\mathcal{D}$ and $\mathcal{D}'$.

A particularly useful inequality that is straight-forward to see is:
$$ | \epsilon_{S}(h, h') - \epsilon_{T}(h, h') | \leq \frac{1}{2} d_{\mathcal{H}\Delta\mathcal{H}}(\mathcal{D}_S, \mathcal{D}_T) $$
which can be directly derived from the definition of the divergence. As we should see in the proof section, this is the key trick used to derive a useful bound.

A more general setting is to extend the 0-1 loss to any loss function $\mathcal{L}$, such as the 0-1 $\epsilon$ defined above. In [16], a similar (to the one in definition 1) but different distribution divergence is proposed:

\textbf{Definition 2} {\itshape For a hypothesis space $\mathcal{H}$ for instance space $\mathcal{X}$ to the output space $\mathcal{Y}$, and let $\mathcal{L}: \mathcal{Y} \times \mathcal{Y} \rightarrow \mathbb{R}_{+} $ defines a loss function over $\mathcal{Y}$. The discrepancy distance $\disc_{\mathcal{L}}$ between two distributions $Q_1$ and $Q_2$ over $\mathcal{X}$ is defined by}
\begin{equation}
\disc_{\mathcal{L}}(Q_1, Q_2) = \max_{h, h' \in \mathcal{H}} \Big| \mathcal{L}_{Q_1}(h, h') - \mathcal{L}_{Q_2}(h, h') \Big|
\end{equation}

Notice that this distance is clearly symmetric and it verifies the triangle inequality. In addition to this distribution divergence, for any such loss function, we can also define the leave one out (LOO) risk for a given training dataset $S$ and a supervised learning algorithm $A$ as:
$$ \hat{\mathcal{L}}^{loo}(A, S) = \frac{1}{m} \sum_{i=1}^{m} l(f_{S^{\setminus i}}, (x_i, y_i))  $$
where $l$ is the application of $\mathcal{L}$ on each single sample. This will become crucial in stability approach.

\section{Key Results}

In this report, we focus our survey on three different theoretical bounds that tries to analyze the effectiveness of transfer learning algorithm: one based on VC-theory, one based on Rademacher complexity and one based on algorithm stability. We first present their main results separately, followed by a proof sketch in the next section. Finally, a discussion about their connections and future works will be presented towards the end of the report.

\subsection{VC-theory approach}

We first start with the model complexity approach using the VC-theory to bound the expected risk in the target domain using the empirical/expected risk in the source domain. This is among the earliest work to derive any bound for the transfer learning/domain adaptation problem. The main idea is to utilize the distribution divergence measure defined in definition 1. 

Using this definition, we can obtain the following theorem of generalization bound for domain adaptation from [14] and [15]:

\textbf{Lemma 1} {\itshape Let $\mathcal{H}$ be a hypothesis space of VC-dimension d, the following holds with probability at least $1 - \delta$ (over the choice of the samples), for every $h \in \mathcal{H}$,}
\begin{equation}\label{eq:3}
\epsilon_{T}(h) \leq \epsilon_{S}(h) + \frac{1}{2} d_{\mathcal{H}\Delta\mathcal{H}}(\mathcal{D}_S, \mathcal{D}_T) + \lambda
\end{equation}
{\itshape where $\lambda = \min\limits_{h \in \mathcal{H}} \epsilon_{S}(h) + \epsilon_{T}(h)$ is the combined risk of the ideal hypothesis.}

Using this lemma, as we will see later in the next section, we can obtain the following main theorem of the VC-theory approach.

\textbf{Theorem 1} {\itshape Let $\mathcal{H}$ be a hypothesis space of VC-dimension d and $\mathcal{U}_S$, $\mathcal{U}_T$ be unlabeled samples of size $m'$ each, drawn from $\mathcal{D}_S$ and $\mathcal{D}_T$, respectively. Let $\hat{d}_{\mathcal{H}\Delta\mathcal{H}}(\mathcal{U}_S, \mathcal{U}_T)$ be the empirical distance on $\mathcal{D}_S$ and $\mathcal{D}_T$, induced by the symmetric difference hypothesis space and $\hat{\epsilon}_S(h)$ be the empirical risk. With probability at least $1 - \delta$ (over the choice of the samples), for every $h \in \mathcal{H}$,}
\begin{equation}\label{eq:4}
\epsilon_{T}(h) \leq \hat{\epsilon}_{S}(h) + \sqrt{\frac{4(d\log(\frac{2em}{d})+\log(\frac{4}{\delta}))}{m}} + \frac{1}{2} \hat{d}_{\mathcal{H}\Delta\mathcal{H}}(\mathcal{U}_S, \mathcal{U}_T) + 4\sqrt{\frac{2d\log(2m')+\log(\frac{4}{\delta})}{m'}} + \lambda
\end{equation}
{\itshape where $m$ is the size of the training labeled samples, drawn from $\mathcal{D}_T$.}

Now we try to interpret this bound. First notice that the bound is relative to $\lambda$ since if there is hypothesis that can perform well on both domains, there is no hope for successful transfer learning. Other than that, the bound consists of an empirical measure of risk in the source domain and an empirical measure of the divergence between the source and the target plus two extra terms that are pretty standard in VC bounds. Basically, this bound that suggests that if $\lambda$ is small, i.e. there exists at least some hypothesis that can do well in both the source domain and the target domain, then the performance in the target domain is upper bounded by the performance within the source domain plus the divergence between their distributions. Although this is rather intuitive, this is nonetheless the first formal proof of that intuition. 

Although the bound in Theorem 1 shows some guarantee of the domain adaptation algorithm, perhaps a more useful case is where we not only try to minimize the empirical risk in the source domain but also minimize the risk in the target domain directly. This is can be achieved by introducing a hyper-parameter $\alpha$ and minimize the following risk:
$$ \hat{\epsilon}_{\alpha}(h) = \alpha \hat{\epsilon}_{T}(h) + (1 -\alpha) \hat{\epsilon}_{S}(h) $$

Then, [15] further proofs the following theorem.

\textbf{Theorem 2} {\itshape Let $\mathcal{H}$ be a hypothesis space of VC-dimension d and $\mathcal{U}_S$, $\mathcal{U}_T$ be unlabeled samples of size $m'$ each, drawn from $\mathcal{D}_S$ and $\mathcal{D}_T$, respectively. Let $\hat{d}_{\mathcal{H}\Delta\mathcal{H}}(\mathcal{U}_S, \mathcal{U}_T)$ be the empirical distance on $\mathcal{D}_S$ and $\mathcal{D}_T$, induced by the symmetric difference hypothesis space. Let $S$ be a labeled sample of size $m$ generated by drawing $\beta m$ points from $\mathcal{D}_T$ and $(1 - \beta) m$ points from $\mathcal{D}_S$, labeling them according to $f_S$ and $f_T$, respectively. If $\hat{h} \in \mathcal{H}$ is the empirical minimizer of $\hat{\epsilon}_{\alpha}(h)$ on S and $h^*_T = \min_{h\in\mathcal{H}} \epsilon_T(h)$ is the target risk minimizer, then with probability at least $1 - \delta$ (over the choice of the samples),}
\begin{equation}\label{eq:5}
\begin{split}
\epsilon_{T}(\hat{h}) & \leq \epsilon_{T}(h^*_T) + 2 \sqrt{\frac{\alpha^2}{\beta}+\frac{(1-\alpha)^2}{1-\beta}}\sqrt{\frac{d\log(2m)-\log(\delta)}{2m}} + \\
& 2(1-\alpha)(\frac{1}{2} \hat{d}_{\mathcal{H}\Delta\mathcal{H}}(\mathcal{U}_S, \mathcal{U}_T) + 4\sqrt{\frac{2d\log(2m')+\log(\frac{4}{\delta})}{m'}} + \lambda)
\end{split}
\end{equation}

This bound is different from the one in theorem 1 in that it is bounded by the theoretical minimizer rather than the empirical risk measurement. However, we can see that setting $\alpha = 0$ amounts to the same setting in theorem 1 which ignores the target labeled data completely. Similarly setting $\alpha = 1$ would ignore the source data completely. Therefore, it is crucial to balance the trade-off controlled by $\alpha$. At the optimal $\alpha$ which minimizes the right hand side, the bound is always at least as either of these two settings. In practice, typically we have much more source labeled data than the target data and therefore this $\alpha$ should be biased towards the source.

Finally, we consider a more realistic case of transfer learning with multiple sources. We will examine algorithms that minimize convex combinations of training errors over the labeled examples from each source domain. Suppose we are now given $N$ source domains and let $m_j = \beta_j m$ with $\sum_{j=1}^N \beta_j = 1$ where $m$ is the size of total labeled data in all source domains. Let the non-negative vector $\alpha$ denotes the domain weights as $\sum_{j=1}^N \alpha_j = 1$, we define the empirical $\alpha$-weighted error of function $h$ as:
$$ \hat{\epsilon}_{\alpha}(h) = \sum_{j=1}^N \alpha_j \hat{\epsilon}_j(h) $$
and similar to the optimal risk $\lambda$ defined in theorem 1, we define the error of the multi-source ideal hypothesis of weight defined by $\alpha$ as:
$$ \lambda_{\alpha} = \min_{h\in\mathcal{H}} \{ \epsilon_T(h) + \sum_{j=1}^N \alpha_j \epsilon_j(h) \} $$
Then we have the following theorem [15].

\textbf{Theorem 3} {\itshape Let $\mathcal{H}$ be a hypothesis space of VC-dimension d and suppose we are given $m_j$ labeled instances from source $S_j$ for j = 1,..,N. For a fixed vector of weights $\alpha$, let $\hat{h} = \arg\min_{h\in\mathcal{H}} \hat{\epsilon}_{\alpha}(h)$, and let $h^*_T = \arg\min_{h\in\mathcal{H}} \epsilon_T(h)$. Then for any $\delta \in (0, 1)$, with probability at least $1 - \delta$ (over the choice of the samples),}
\begin{equation}\label{eq:6}
\epsilon_{T}(\hat{h}) \leq \epsilon_{T}(h^*_T) + 2 \sqrt{\sum_{j=1}^N \frac{\alpha_j^2}{\beta_j}} \sqrt{\frac{d\log(2m)-\log(\delta)}{2m}} + 2(\frac{1}{2} d_{\mathcal{H}\Delta\mathcal{H}}(\mathcal{D}_{\alpha}, \mathcal{D}_T) + \lambda_{\alpha})
\end{equation}

This bound is the first theoretical generalization bound for multi-source transfer learning problem. One can replace the divergence term with its empirical measure. However, this bound actually is not quite useful as it tells us nothing except that a more even weight of $\alpha$ and $\beta$ should be preferred. \textbf{We prove a more useful bound in section 3.4 and we will delay the comparison there.}

\subsection{Rademacher complexity approach}

The second approach of the problem used the Rademacher complexity. For the convenience of readers, we first rewrite the definition of Rademacher complexity here.

\textbf{Definition 3} {\itshape Let $H$ be a set of real-valued functions defined over a set $X$. Given a sample $X \in X^m$, the empirical Rademacher complexity of $H$ is defined as follows:}
\begin{equation}
\hat{\mathcal{R}}_S(H) = \frac{2}{m} \mathbb{E}_{\sigma}[\sup_{h\in H} |\sum_{i=1}^{m} \sigma_i h(x_i)| \Big| S = (x_1,...,x_m)].
\end{equation}
{\itshape The expectation is taken over $\sigma$ where $\sigma_i$ are independent uniform random variables taking values in $\{-1, +1\}$. The Rademacher complexity of a hypothesis set $H$ is defined as the expectation of $\hat{\mathcal{R}}_S(H)$ over all samples of size m:}
\begin{equation}
\mathcal{R}_m(H) = \mathbb{E}_{S}[\hat{\mathcal{R}}_S(H) \Big| |S| = m].
\end{equation}

Notice that this distance is clearly symmetric and it verifies the triangle inequality.

The following results using the above definitions are mainly from [16]. Their approach is similar to the one presented in section 3.1 using the VC-theory but using a different divergence measure and Rademacher complexity. To start with, they first prove a very similar bound to theorem 1.

\textbf{Theorem 4} {\itshape Assume that the loss function $\mathcal{L}$ is symmetric and obeys the triangle inequality. Then, for any hypothesis $h \in \mathcal{H}$, the following holds}
\begin{equation}
\mathcal{L}_{T}(h, f_T) \leq \mathcal{L}_{T}(h_T^*, f_T) + \mathcal{L}_{S}(h, h_S^*) + \disc(S, T) + \mathcal{L}_{S}(h_S^*, h_T^*)   
\end{equation}
{\itshape where $h_S^* = \arg\min_{h\in\mathcal{H}} \mathcal{L}_S(h, f_S)$ is the minimizer and similarly for $h_T^*$.}

Similar to theorem 1, this bound gives an upper bound on a function in the target domain using the true minimizer in the target and the source domain. While all terms are theoretical, it clearly leads to bounds based on the empirical error of $h$ on a sample drawn according to the source domain and/or the target domain. Then, the main result of [16] is the following extension of theorem 4, with the application of Rademacher classification bound applied on the distribution divergence (as in Corollary 7 in [16]).

\textbf{Theorem 5} {\itshape Let $\mathcal{H}$ be a family of functions mapping $\mathcal{X}$ to $\{0, 1\}$. Let $S$ be the source domain and $\hat{S}$ be the corresponding empirical distribution for a sample $\mathcal{S}$, and let $T$ be the target domain and $\hat{P}$ be the corresponding empirical distribution for a sample $\mathcal{T}$. Then, for any $h \in \mathcal{H}$, with probability at least $1-\delta$, with $\mathcal{S}$ of size m and $\mathcal{T}$ of size n, the following generalization bound holds for the 0-1 loss:}
\begin{equation}
\begin{split}
\mathcal{L}_{T}(h, f_T) - \mathcal{L}_T(h_T^*, f_T) & \leq \mathcal{L}_{\hat{S}}(h, h_S^*) + \disc_{\mathcal{L}}(\hat{S}, \hat{T}) + (4+\frac{1}{2}) \hat{\mathcal{R}}_{\mathcal{S}}(\mathcal{H}) + \\ & 4 \hat{\mathcal{R}}_{\mathcal{T}}(\mathcal{H}) + 4 \sqrt{\frac{\log(\frac{8}{\delta})}{2m}} + 3 \sqrt{\frac{\log(\frac{8}{\delta})}{2n}} + \mathcal{L}_{S}(h_S^*, h_T^*)
\end{split}
\end{equation}
{\itshape where $h_S^* = \arg\min_{h\in\mathcal{H}} \mathcal{L}_S(h, f_S)$ is the minimizer and similarly for $h_T^*$.}

As we will discuss later in the last section, this bound is similar to theorem 1 but it is also in general tighter. This is by far the best result of domain adaptation generalization bound using model complexity approach. 

\subsection{Stability approach}

Finally, we introduce the stability approach of deriving learning bounds for transfer learning algorithms. However, this approach is based on a very special approach to transfer learning named hypothesis transfer learning. It assumes no direct access to the training data in the source domain. Instead, one or more learned functions from the source domain are available. There are two key benefits for such approach. One it relieves the requirement of large data storage and two the distribution divergence is no longer a problem in deriving the complexity bound (to some degree). In particular, [18] derives the learning bound for the following transfer learning algorithm.

\textbf{Algorithm 1} {\itshape An regularized least squares algorithm. Assume that a hypothesis $f'$ is trained from the source domain, then we define a new hypothesis for the target domain as:}
\begin{equation}
f(x) = T_C(x^T\hat{w}) + f'(x),
\end{equation}
where
\begin{equation}
\hat{w} := \arg\min_u \frac{1}{m} \sum_{i=1}^m (u^Tx_i -y_i+f'(x_i))^2 + \lambda \Vert u \Vert^2
\end{equation}
and the truncation function $T_C(y)$ is defined as $T_C(y)=\min(\max(y,-C),C)$.

Then, authors of [18] show that, using the hypothesis stability definition and theory derived in [11] (not included due to space limits), the following learning bound:

\textbf{Theorem 6} {\itshape Set $\lambda \geq \frac{1}{m}$. If $C \geq B+\Vert f' \Vert_{\infty}$, then for Algorithm 1 we have}
\begin{equation}
\mathbb{E}[(\mathcal{L}_T(f) - \hat{\mathcal{L}}^{loo}(f))^2] = O(C^2 \frac{\sqrt{\mathcal{L}_T(f')T_{C^2}(\frac{\mathcal{L}_T(f')}{\lambda})+\mathcal{L}_T(f')^2}}{m\lambda^{1.5}})
\end{equation}
{\itshape If $C = \infty$, then for Algorithm 1 we have}
\begin{equation}
\mathbb{E}[(\mathcal{L}_T(f) - \hat{\mathcal{L}}^{loo}(f))^2] = O(\frac{\mathcal{L}_T(f')(\Vert f' \Vert_{\infty}+B)^2}{m\lambda^3})
\end{equation}

\subsection{Improved bound for multiple sources}

Here, we prove a novel bound based on the VC-theory used in Theorem 1. The trick is to use labeled data from all source to help evaluate the stability of the learned hypothesis.

\textbf{Theorem 7} {\itshape Let $\mathcal{H}$ be a hypothesis space of VC-dimension d and $\hat{d}_{\mathcal{H}\Delta\mathcal{H}}(S_i^U, T)$ be the empirical distributional distance between the $i^{th}$ source and the target domain, induced by the symmetric difference hypothesis space. Then, for any $\mu \in (0, 1)$ and any $\hat{h} = \sum_i \alpha_i \hat{h}_i$ where $\hat{h}_i \in \mathcal{H}$ and $\sum \alpha_i = 1$, the following holds with probability at least $1 - \delta$, }
\begin{equation}
\small
\begin{split}
 \epsilon_T(\hat{h}) & \leq \Bigg[\sum\limits_i^K \alpha_i \bigg[ \mu \Big[\hat{\epsilon}_{S_i}(\hat{h}_i) + \frac{1}{2} \hat{d}_{\mathcal{H}\Delta\mathcal{H}}(S_i^U, T)\Big] \\
 & + \frac{1-\mu}{K-1} \sum\limits_{j\neq i}^K \Big[\hat{\epsilon}_{S_j}(\hat{h}_i) + \frac{1}{2} \hat{d}_{\mathcal{H}\Delta\mathcal{H}}(S_j^U, T) \Big] \Bigg] \\  
& + \bigg(\sum\limits_i^K \alpha_i \sqrt{\frac{\mu^2}{\beta_i} + (\frac{1-\mu}{K-1})^2 \sum_{j\neq i}\frac{1}{\beta_j}}\bigg)\sqrt{\frac{d\log(2m)-\log(\delta)}{2m}} \bigg] \\
& + 4 \sqrt{\frac{2d \log(2m')+\log(\frac{4}{\delta})}{m'}} + \lambda_{\alpha, \mu},
\end{split}
\end{equation}
\normalsize 
 {\itshape where $\epsilon_{*}(h)$ is the expected risk of h in the corresponding domain, $m=\sum_i^K n_i^L$ is the sum of labeled sizes in all sources, $\beta_i = n_i^L/m$ is the ratio of labeled data in the $i^{th}$ source, and $\lambda_{\alpha, \mu}$ is the risk of the ideal multi-source hypothesis weighted by $\alpha$ and $\mu$.}

\section{Proof Outlines}

Due to space limits, we are only presenting here the proof of the bound that we derived in Theorem 7. First we show that:
{\footnotesize
\begin{align*} 
& \quad \sum_i^K \alpha_i \Big[\mu\lambda_i + \frac{1-\mu}{K-1} \sum_{j\neq i} \lambda_j \Big]\\
& = \sum_i^K \alpha_i \Big[\mu(\epsilon_{T}(h^*_i)+\epsilon_{S_i}(h^*_i)) + \frac{1-\mu}{K-1} \sum_{j\neq i} (\epsilon_{T}(h^*_j)+\epsilon_{S_j}(h^*_j)) \Big]\\
& \leq \sum_i^K \alpha_i \Big[\mu(\epsilon_{T}(h_{\alpha, \mu}^*)+\epsilon_{S_i}(h_{\alpha, \mu}^*)) + \frac{1-\mu}{K-1} \sum_{j\neq i} (\epsilon_{T}(h_{\alpha, \mu}^*)+\epsilon_{S_j}(h_{\alpha, \mu}^*)) \Big]\\
& = \epsilon_{T}(h_{\alpha, \mu}^*)+\sum\limits_i^K (\alpha_i \mu + (1-\alpha_i)\frac{1-\mu}{K-1})\epsilon_{S_i}(h_{\alpha, \mu}^*) = \lambda_{\alpha, \mu}
\end{align*}
}%
where $\lambda_{\alpha, \mu} =  \min_{h\in \mathcal{H}}\epsilon_{T}(h)+\sum\limits_i^K (\alpha_i \mu + (1-\alpha_i)\frac{1-\mu}{K-1})\epsilon_{S_i}(h)$ and $h_{\alpha, \mu}^* = \mathop{\arg\min}\limits_{h\in\mathcal{H}} \epsilon_{T}(h)+\sum\limits_i^K (\alpha_i \mu + (1-\alpha_i)\frac{1-\mu}{K-1})\epsilon_{S_i}(h)$. 

Then, using the above inequality we can prove the bound as:
{\scriptsize
\begin{align*} 
\epsilon_T(\hat{h}) &= \epsilon_T(\sum\limits_i^k \alpha_i \hat{h_i}) = \sum\limits_i^k \alpha_i \epsilon_T(\hat{h_i}) = \sum\limits_i^k \alpha_i \Big[ \mu \epsilon_T(\hat{h_i}) +  \frac{1-\mu}{k-1} \sum\limits_{j\neq i}^k \epsilon_T(\hat{h_i}) \Big]\\ 
 &\leq  \sum\limits_i^k \alpha_i \Bigg[ \mu \Big[\epsilon_{S_i}(\hat{h_i}) + \frac{1}{2} \hat{d}_{\mathcal{H}\bigtriangleup\mathcal{H}}(U_{S_i}, U_T) + 4 \sqrt{\frac{2d \log(2m')+\log(\frac{4}{\delta})}{m'}} + \lambda_i \Big] + \\
 & \qquad\qquad \frac{1-\mu}{k-1} \sum\limits_{j\neq i}^k \Big[\epsilon_{S_j}(\hat{h_i}) + \frac{1}{2} \hat{d}_{\mathcal{H}\bigtriangleup\mathcal{H}}(U_{S_j}, U_T) + 4 \sqrt{\frac{2d \log(2m')+\log(\frac{4}{\delta})}{m'}} + \lambda_j \Big] \Bigg] \text{(Theorem 1 of [15])}\\
 &\leq  \Bigg[\sum\limits_i^k \alpha_i \bigg[ \mu \Big[\epsilon_{S_i}(\hat{h_i}) + \frac{1}{2} \hat{d}_{\mathcal{H}\bigtriangleup\mathcal{H}}(U_{S_i}, U_T)\Big] + \frac{1-\mu}{k-1} \sum\limits_{j\neq i}^k \Big[\epsilon_{S_j}(\hat{h_i}) + \frac{1}{2} \hat{d}_{\mathcal{H}\bigtriangleup\mathcal{H}}(U_{S_j}, U_T) \Big] \bigg]\Bigg] + \\
 & \qquad\qquad 4 \sqrt{\frac{2d \log(2m')+\log(\frac{4}{\delta})}{m'}} + \lambda_{\alpha, \mu} \\
  &\leq  \Bigg[\sum\limits_i^k \alpha_i \bigg[ \mu \Big[\hat{\epsilon}_{S_i}(\hat{h_i}) + \frac{1}{2} \hat{d}_{\mathcal{H}\bigtriangleup\mathcal{H}}(U_{S_i}, U_T)\Big] + \frac{1-\mu}{k-1} \sum\limits_{j\neq i}^k \Big[\hat{\epsilon}_{S_j}(\hat{h_i}) + \frac{1}{2} \hat{d}_{\mathcal{H}\bigtriangleup\mathcal{H}}(U_{S_j}, U_T) \Big] + \\
  & \qquad\qquad \sqrt{\frac{\mu^2}{\beta_i} + (\frac{1-\mu}{k-1})^2 \sum_{j\neq i}\frac{1}{\beta_j}}\sqrt{\frac{d\log(2m)-\log(\delta)}{2m}} \bigg]\Bigg] + 4 \sqrt{\frac{2d \log(2m')+\log(\frac{4}{\delta})}{m'}} + \lambda_{\alpha, \mu} \text{(Lemma 4 of [15])} \\
&=  \Bigg[\sum\limits_i^k \alpha_i \bigg[ \mu \Big[\hat{\epsilon}_{S_i}(\hat{h_i}) + \frac{1}{2} \hat{d}_{\mathcal{H}\bigtriangleup\mathcal{H}}(U_{S_i}, U_T)\Big] + \frac{1-\mu}{k-1} \sum\limits_{j\neq i}^k \Big[\hat{\epsilon}_{S_j}(\hat{h_i}) + \frac{1}{2} \hat{d}_{\mathcal{H}\bigtriangleup\mathcal{H}}(U_{S_j}, U_T) \Big] \Bigg] + \\  
& \qquad\qquad  \bigg(\sum\limits_i^k \alpha_i \sqrt{\frac{\mu^2}{\beta_i} + (\frac{1-\mu}{k-1})^2 \sum_{j\neq i}\frac{1}{\beta_j}}\bigg)\sqrt{\frac{d\log(2m)-\log(\delta)}{2m}} \bigg]+ 4 \sqrt{\frac{2d \log(2m')+\log(\frac{4}{\delta})}{m'}} + \lambda_{\alpha, \mu}
\end{align*}

}%

\section{Conclusion}
In this report, we present three different approaches to deriving learning bounds of transfer learning problem, one using VC-theory, one using Rademacher complexity and one using algorithmic stability.

For the two model complexity approaches in Theorem 1 and Theorem 5, we can see that both of them require some form of distribution divergence measure and the risk bound is upper bounded by this divergence. It is consistent with our intuition that there is no hope for successful transfer if the divergence between the source and the target is too large. They also both contain a expected risk term of the ideal hypothesis, which also shows that such best possible performance is also needed for low risk in the target domain. However, when comparing the differences between these two bounds, one can observe that there are three error terms involving the target function in Theorem 1 while there is only one in Theorem 5. In the extreme case, one can see that the bound in Theorem 1 might become vacuous for moderate values of expected target function risk and a factor of 3 might arise when comparing the two bounds. While in general the two bounds are incomparable, we can still say that Theorem 5 might be a more realistically useful bound. On the other hand, the stability bound in Theorem 6 is rather experimental than practical. It kinds of skipping the distribution divergence problem by only requiring the hypothesis trained from the source so it is not really showing a learning bound of "transfer learning" in that sense. However, it is still interesting to see a different approach which gives us some insights on how leave one out error can be more useful than other empirical error measure.

In terms of the multi-source transfer learning case, the bound in Theorem 3 does not tell us much. For the novel work we did in Theorem 7, by introducing a concentration factor $\mu$, we replace the size of labeled data for each source with the total size $m$ of all sources, resulting in a tighter bound. Suppose that the hypothesis $\hat{h}_i$ is learned using data in the $i^{th}$ source, then the bound suggests to use peers to evaluate the reliability of $\hat{h}_i$ while the optimal weights should consider both proximity and reliability of the $i^{th}$ source by controlling the value of $\mu$.

Finally, we conclude that while some great works have been done in the theoretical analysis of transfer learning, the theory is still minimal compared to the practical success of the field. In future, a very interesting direction would be deriving learning bounds for deep learning and further combining that with transfer learning, as many transfer learning methods are based on that.

\newpage

\section*{References}

\medskip

\small

[1] Pan, Sinno Jialin and Yang, Qiang.  A survey on transfer learning. {\it IEEE  Transactions  on  knowledge  and  data engineering}, 22(10):1345–1359, 2010.

[2] Raina, Rajat, Battle, Alexis, Lee, Honglak, Packer, Ben-jamin, and Ng, Andrew Y. Self-taught learning: transferlearning from unlabeled data. {\it International Confer-ence on Machine learning (ICML)}, pp. 759–766, 2007.

[3] Pan, Sinno Jialin, Tsang, Ivor W, Kwok, James T, and Yang,Qiang. Domain adaptation via transfer component analysis.{\it IEEE Transactions on Neural Networks}, 22(2):199–210, 2011.

[4] Zhang,  Lei,  Zuo,  Wangmeng,  and Zhang,  David. Lsdt:Latent sparse domain transfer learning for visual adaptation. {\it IEEE Transactions on Image Processing}, 25(3):1177–1191, 2016.

[5] Blitzer, John, Dredze, Mark, Pereira, Fernando, et al.  Biographies,  bollywood,  boom-boxes and blenders:  Do-main adaptation for sentiment classification. {\it Meeting of the Association of Computational Linguistics (ACL)},pp. 440–447, 2007.

[6] Long, Mingsheng, Wang, Jianmin, and Jordan, Michael I.Deep transfer learning with joint adaptation networks. {\it International Conference on Machine Learning (ICML)},pp. 2208–2217, 2017.

[7] Moon, Seungwhan and Carbonell, Jaime. Completely heterogeneous  transfer  learning  with  attention-what  and what not to transfer. {\it International Joint Conference on Artificial Intelligence (IJCAI)}, pp. 2508–2514, 2017.

[8] Wei, Ying, Zheng, Yu, and Yang, Qiang.  Transfer knowledge between cities. In KDD, pp. 1905–1914, 2016.

[9] Mohri, M., Rostamizadeh, A., and Talwalkar, A. {\it Foundations of machine learning}. MIT press, 2012.

[10] Vapnik, Vladimir, Levin, Esther and Cun, Yann Le, Measuring the VC-dimension of a learning machine, {\it Neural computation}, MIT press, 1994.

[11] Bousquet, Olivier and Elisseeff, Andre, Stability and generalization, {\it Journal of machine learning research}, 2002.

[12] Gong, Boqing and Shi, Yuan and Sha, Fei and Grauman, Kristen, Geodesic flow kernel for unsupervised domain adaptation, {\it Computer Vision and Pattern Recognition (CVPR), 2012 IEEE Conference on}, 2012.

[13] Ben-David, Shai, Blitzer, John, Crammer, Koby, and Pereira,Fernando. Analysis of representations for domain adaptation.  {\it Advances in Neural Information Processing Systems (NIPS)}, pp. 137–144, 2007.

[14] Ben-David, Shai, Blitzer, John, Crammer, Koby, Kulesza,Alex, Pereira, Fernando, and Vaughan, Jennifer Wortman. A theory of learning from different domains. {\it Machine learning}, 79(1):151–175, 2010.

[15] Blitzer, John, Crammer, Koby, Kulesza, Alex, Pereira, Fer-nando, and Wortman, Jennifer. Learning bounds for domain  adaptation.  {\it Advances  in  Neural  Information Processing Systems (NIPS)}, pp. 129–136, 2008.

[16] Mansour, Yishay, Mohri, Mehryar, and Rostamizadeh, Af-shin. Domain adaptation: learning bounds and algorithms.{\it Conference On Learning Theory (COLT)}, 2009.

[17] Fei-Fei, L., Fergus, R., and Perona, P. One-shot learning of object categories. {\it Pattern Analysis and Machine Intelligence}, 2006.

[18] Kuzborskij, Ilja and Orabona, Francesco, Stability and hypothesis transfer learning. {\it International Conference on Machine Learning (ICML)}, 2013.

[19] Rosenstein, Michael T, Marx, Zvika, Kaelbling, Leslie Pack,and Dietterich, Thomas G. To transfer or not to transfer.{\it NIPS-2005 Workshop on Inductive Transfer: 10 Years Later}, 2005.

[20] Mansour, Yishay, Mohri, Mehryar, and Rostamizadeh, Af-shin.  Domain adaptation with multiple sources.  {\it Advances in Neural Information Processing Systems (NIPS)},pp. 1041–1048, 2009.

[21] Du, S., Koushik, J., Singh, A. and Poczos, B., Hypothesis transfer learning via transformation functions, {\it Advances in Neural Information Processing Systems (NIPS)}, 2017.

\end{document}